\documentclass[10pt,twocolumn,letterpaper]{article}

\usepackage{iccv}      % To produce the REVIEW version

\usepackage[pagebackref,breaklinks,colorlinks]{hyperref}
\usepackage{multirow} 
\usepackage{algorithm}
\usepackage{algorithmicx}
\usepackage{algpseudocode}
\usepackage{stfloats}
\usepackage{graphicx}
\usepackage{xcolor}
\usepackage{comment}
\usepackage{amsmath}
\usepackage[bottom]{footmisc}
\usepackage{subfloat}
\usepackage{tabularx}
\usepackage{stfloats}
\usepackage{color}

\usepackage{tcolorbox}
\tcbuselibrary{breakable}
\definecolor{my_green}{RGB}{40,154,121}
\definecolor{my_red}{RGB}{176,46,46}
\definecolor{iccvblue}{rgb}{0.21,0.49,0.74}

\def\paperID{10683} % *** Enter the Paper ID here
\def\confName{ICCV}
\def\confYear{2025}

\title{Zero-Shot Subject-Centric Generation for Creative Application Using Entropy Fusion}

\author{
Kaifeng Zou$^{1*}$ \quad Xiaoyi Feng$^{2*}$ \quad Peng Wang$^{2}$ \quad Tao Huang$^{3}$ \quad \\
Zizhou Huang$^{1}$ \quad Zhang Haihang$^{1}$ \quad Yuntao Zou$^{2}$ \quad Dagang Li$^{2\dagger}$ \\
\small{$^1$ Link-To, Shenzhen, China \quad $^2$ Macau University of Science and Technology, Macau, China} \\ \small{$^3$ Huazhong University of Science and Technology, Wuhan, China} \\
\small{$^*$ Equal contribution \quad $^\dagger$ Corresponding author}
}

\begin{document}
% \pgfplotsset{compat=1.14}
%\maketitle
% \setcounter{page}{1}
% \maketitlesupplementary
% \label{appendix}
% \input{sec/sup/sup1}
% \clearpage
% \input{sec/sup/sup2}

\twocolumn[{
\renewcommand\twocolumn[1][]{#1}
\maketitle
\begin{center}
    \captionsetup{type=figure}
    \includegraphics[width=0.95\textwidth]{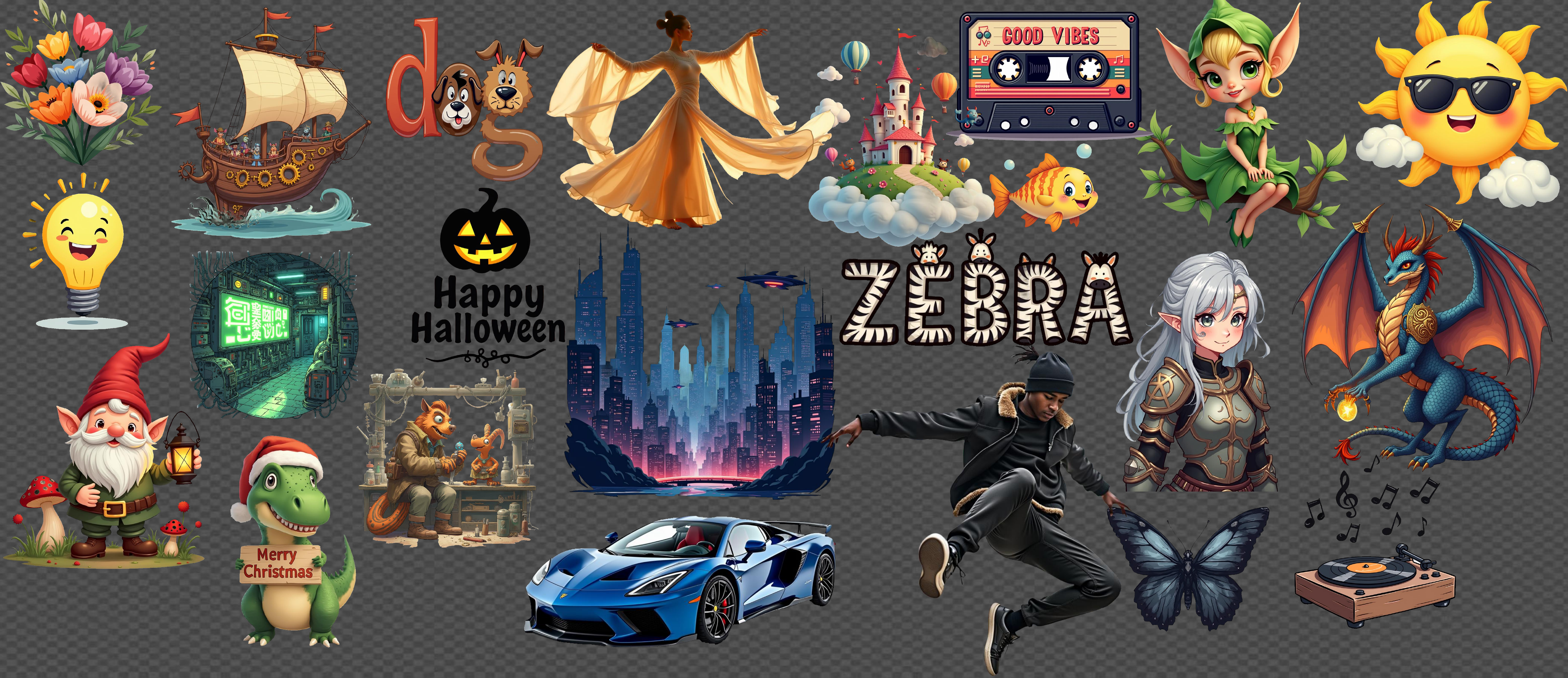}
    \vspace{-5pt}
    \captionof{figure}{Subject-centric images generated by our methods.}
    \vspace{10pt}
\end{center}
}]

\begin{abstract}
Generative models are widely used in visual content creation. However, current text-to-image models often face challenges in practical applications—such as textile pattern design and meme generation—due to the presence of unwanted elements that are difficult to separate with existing methods. Meanwhile, subject-reference generation has emerged as a key research trend, highlighting the need for techniques that can produce clean, high-quality subject images while effectively removing extraneous components.
To address this challenge, we introduce a framework for reliable subject-centric image generation. In this work, we propose an entropy-based feature-weighted fusion method to merge the informative cross-attention features obtained from each sampling step of the pretrained text-to-image model FLUX, 
enabling a precise mask prediction and subject-centric generation.
Additionally, we have developed an agent framework based on Large Language Models (LLMs) that translates users' casual inputs into more descriptive prompts, leading to highly detailed image generation. Simultaneously, the agents extract primary elements of prompts to guide the entropy-based feature fusion, ensuring focused primary element generation without extraneous components.
Experimental results and user studies demonstrate our methods generates high-quality subject-centric images, outperform existing methods or other possible pipelines, highlighting the effectiveness of our approach.
\end{abstract}

\section{Introduction}
\label{intro}
%In the current digital era, visual elements play a crucial role in the field of design.
%With the advancement of artificial intelligence technology, particularly in the field of image generation, an increasing amount of research is focusing on creating high-quality, highly realistic visual elements \cite{latentdiffusion, podell2024sdxl, latentdiffusion, rectified_flow, imagen, chen2024pixartalpha, BlackForest2024}.
%However, these technologies still face significant challenges in meeting design professionals' practical needs.
Object-level image customization \cite{chen2024anydoor, sarukkai2024collage, liu2023customizable, zhang2024ssr} and video customization \cite{wang2024customvideo, huang2024subjectdrive} have become increasingly popular, providing generative models with more versatile applications. In particular, Vidu \cite{vidu} and PIKA\cite{pika} have introduced multi-subject reference video generation models, which have had a significant impact on social media.  Thus, there is growing interest in subject-centric generation, where models focus solely on generating the primary subject while excluding background or contextual elements.

%Moreover, in real-world design workflows – whether creating brand visuals, marketing materials, or presentation graphics – designers often require specific elements to serve as editable design assets. These may include custom patterns for textile printing, modular poster components, reusable PPT graphical elements, or brand identity components like company logos. 

Current AI-generated models \cite{latentdiffusion, podell2024sdxl, rectified_flow, imagen, chen2024pixartalpha, BlackForest2024} can produce visually impressive results. However, they often lack structural decomposition and layer-wise editability, which are essential for seamless integration into professional design workflows.
The core challenge lies in bridging the gap between image generation and real visual-content-creation workflow requirements. Users need AI tools that not only produce high-quality outputs, but also support element-level control, flexible recomposition, and natural blending with existing design contexts.

%Similarly, most visual design software, such as PhotoShop \cite{photoshop} relies on images with transparency channels to enable users to create content and compositions freely. 
%Large-scale image generation models \cite{podell2024sdxl, BlackForest2024} have seen widespread adoption and have profoundly affected the visual arts. However, the challenge of layered image generation has not been thoroughly addressed.

One possible solution is RGBA image generation. For example, ~\cite{layerdiffusion} constructed a dataset of one million images with transparency channels and used it to train a diffusion model capable of generating images with transparency. While this method enables the generation of RGBA images that can be applied in certain contexts, it requires a substantial amount of training data. Additionally, the generated images often exhibit continuous alpha channel values, which may not be necessary for many practical applications. 
Through extensive industry surveys, we found that in many practical applications—such as T-shirt prints, stickers, and logos—a continuous alpha channel is not required. Instead, these applications typically need distinct, discrete alpha values for clear separation of objects from the background, making the use of continuous alpha channel values less suitable and potentially introducing unnecessary complexity.

 To bridge this gap, a prevalent solution adopts a two-stage generation-then-segmentation paradigm. In the first stage, advanced image generation models like \cite{BlackForest2024} produce high-quality synthetic images. These outputs are then processed through segmentation modules, where universal segmentation frameworks such as SAM \cite{kirillov2023segment} can parse all image elements, while salient object detectors like \cite{gao2024multi} focus on extracting primary subjects. However, this approach often requires expert knowledge and the integration of multiple models, which can complicate the process and reduce its accessibility.
 
%However the quality of its RGBA generation depends on the nouns of individual words in the sentence and the manual selection of nouns to remove, which makes the prompt highly limited and results in poor attention map quality.

%Moreover, these methods are more inclined to generate RGBA images with a gradient alpha channel, making it difficult to provide a definitive binary alpha map (completely transparent or completely opaque), which complicates the re-integration of the various elements.  When attempting to merge the image generation results of both methods with the selected background, the gradient alpha channel leads to conflicts between foreground and background, resulting in blurriness.

To improve the practical value of AI-generated images in visual content creation, it is essential to address the challenges of subject-centric images generation and ensuring the effective integration of elements. In response to these needs, we propose a novel image generation pipeline that seamlessly integrates with advanced generative models.
In this work, we leverage the state-of-the-art pre-trained text-to-image model, such as FLUX \cite{BlackForest2024}, to generate the visual elements. Many researches has proven that it is effective to condition diffusion model by manipulating cross-attention or self-attention layers \cite{hertz2023prompt, couairon2023diffedit, tumanyan2023plug, cao2023masactrl},
 indicating that cross-attention maps convey valuable information \cite{tang2022daam}. Based on this insight, we propose an entropy-based feature fusion strategy to integrate attention maps linked to the most influential prompt words (typically nouns related to the foreground) during each sampling step. This approach enhances the fidelity of detecting salient element in image generation process by strategically manipulating attention features.
 Moreover, the formulation and selection of prompts have a significant impact on our results. To make the entire process feasible and automated, we design an agent framework, which allows us to obtain richly detailed results by simply input desired elements and style while automatically selecting key elements to guide the subject-centric image generation.

Finally, our method produces faithful image generation results that can be seamlessly adapted to a wide range of industrial applications, from merchandise printing to digital media, ensuring smooth integration with diverse background patterns and ready for real-world production. %outperforming other layered image generation techniques.
%This advantage is attributed to our ability to achieve more accurate alpha channel estimation.
%Furthermore, we also compare our method with the traditional solution for RGBA image generation: generating-then-matting (See~\cref{sec:relatedwork}) \cite{isnet, birefnet} which face the similar challenges as generation methods due to uncertain segmentation map prediction. 

The workflow most similar to ours combines generation with salient object detection (SOD) \cite{borji2019salient}. We have compared our approach with SAM\cite{kirillov2023segment} and its derivatives for SOD\cite{gao2024multi}, as well as with the generation-then-matting method\cite{isnet, birefnet}. Additionally, we conducted comparisons with RGBA-based generation techniques \cite{layerdiffusion, quattrini2024alfie} to assess their performance and suitability for our application.
Extensive experimental results suggest that our method offers greater reliability. This advantage arises from our ability to generate high-quality images while ensuring the precise creation of key elements, indicating that our approach is well-suited for creative applications.

In summary, our contribution can be organized as follows:

\begin{itemize}

    \item  We have developed a zero-shot subject-centric generative approach, which is capable of generating images with accurate synthesis of primary subject, outperforming existing methods.

    \item  We proposed an entropy-based feature fusion module which leverages the informative cross-attention feature map of the diffusion reverse steps, enabling us to accurately retain the chosen elements.
    %enabling us to achieve highly accurate alpha channel.
    
    \item  We designed an agent based on large language models that extends users' simple and casual input into more expressive prompts  while extracting primary elements, filtering out irrelevant objects, greatly enhancing the accuracy of primary elements synthesis.

    %We designed an agent based  on large language models that can generate more expressive keywords from users' simple and casual input,   filtering out irrelevant objects retain the chosen elements
    %simplifies input into a few key elements. The agent automatically expands the prompt and extract keywords,  effectively lowering the usability barrier of the method while 

    \item  Experimental results demonstrate that our approach has achieve SOTA performance in subject-centric generation, and ablation studies show the effectiveness of our designed components. The code will be release publicly to facilitate future research.

\end{itemize}

\section{Related Work}
\label{sec:relatedwork}
\noindent 
\textbf{Diffusion Models.} 
Originated from Diffusion Probabilistic Model \cite{sohl2015deep}, diffusion models began to receive widespread attention when researchers proposed practical training and sampling algorithms \cite{ho2020denoising, song2021denoising, nichol2021improved}.%and showed the effective of diffusion models against GANs in image generation tasks \cite{dhariwal2021diffusion}.
Furthermore, \cite{peebles2023scalable, crowson2024scalable} proposed Diffusion Transformers, which replace the convolutional U-Net structure with transformers, enhancing the scalability and visual quality of diffusion models.
These above works have paved the way for recent large-scale diffusion models like Stable Diffusion \cite{latentdiffusion,  podell2024sdxl} and Flux \cite{BlackForest2024}.

\noindent There are many works which try to enhance the controllability of the generated image from large-scale diffusion models.
One approach is to develop effective fine-tuning methods based on additional guidance information, such as ControlNet \cite{zhang2023adding}, T2I-adapter \cite{mou2024t2i}. 
Another approach is to exploit the zero-shot capabilities of pre-trained diffusion models by proposing tuning-free methods to change the denoising process.
Many of them are based on the manipulation of cross-attention or self-attention layers \cite{hertz2023prompt, couairon2023diffedit, tumanyan2023plug, cao2023masactrl}.  Some in-depth discussions about the role of cross and self-attention layers in stable diffusion are provided in \cite{liu2024towards}.

\noindent 
\textbf{Subject-Centric Generation.} 
Conventional subject-centric creation workflows typically decouple content generation from element extraction. A key challenge in such cascaded frameworks is error accumulation: segmentation accuracy is inherently tied to generation quality, while post-processing techniques, such as segmentation \cite{kirillov2023segment} and SOD\cite{gao2024multi}, often struggle with fine details, leading to imprecise boundaries and missing intricate structures. Additionally, alpha estimation using matting methods \cite{qin2022highly, yao2024vitmatte} can introduce artifacts that are incompatible with standard constraints—particularly the requirement for a binary alpha channel, necessitating an additional thresholding step, which may further degrade edge fidelity.

%Recent efforts attempt tighter integration - LayerDiffusion \cite{li2023layerdiffusion} fine-tunes foundation models for transparency generation, while Alfie \cite{quattrini2024alfie} modifies diffusion attention mechanisms for RGBA output. However, these methods %retain continuous alpha channels ill-suited for binary mask requirements in physical production.
Recent efforts attempt tighter integration: LayerDiffusion \cite{li2023layerdiffusion} fine-tunes foundation models for transparency generation, while Alfie \cite{quattrini2024alfie} modifies diffusion attention mechanisms for RGBA output. However, these methods often struggle with strict alpha compositing constraints, leading to inconsistencies when integrating with standard design pipelines. Additionally, RGBA-based approaches inherently suffer from color bleeding, edge inconsistencies, and difficulty in achieving precise subject-centric generation.

\noindent
\textbf{Prompt Engineering.} With the growing popularity of text-to-image models, researchers focus on the impact of text on the quality of image generation. Some researchers \cite{liu2022design} explore the text design in text-to-image generative models and systematically summarize them into design guidelines to achieve better image quality. 3DALL-E \cite{liu20233dall} tried using GPT-3 to rewrite selected prompts. Prompt-to-prompt \cite{hertz2023prompttoprompt} proposes an editing framework that can semantically edit images by modifying prompts in a pre-trained text-conditional diffusion model. Promptify \cite{brade2023promptify} is based on a large language model and helps users improve original text prompts through a visual interface and interactive system. %Omost \cite{omost} combines a large language model and Tree-of-Thoughts \cite{yao2024tree} to enhance image generation and user-friendliness by optimizing user input prompts and selecting the most appropriate generation model.
Omost \cite{omost} enhance image generation and user-friendliness by optimizing user input prompts and selecting the most appropriate generation model.

Our work is the first attempt to explore prompt optimization by the agent and build an agent framework for prompt expansion and key elements extraction, which can enhance attention map extraction and RGBA image generation.

% 文生图prompt优化：
% https://arxiv.org/abs/2109.06977, https://arxiv.org/abs/2210.11603, https://arxiv.org/abs/2208.01626
% https://arxiv.org/abs/2304.09337, https://github.com/lllyasviel/Omost, 

% 【In this paper, 本文是属于XXX】

\section{Methods}
In this section, we propose a zero-shot method to synthesize images where the primary subject is accurately generated while minimizing the presence of extraneous elements. Additionally, we introduce an agent-based framework that simplifies user input by automatically structuring it into an optimized format for subject generation, reducing the need for manual intervention.
We begin with an introduction of overall framework (\cref{overview}) and then introduce rectified flow and MM-DiT arcitecture (\cref{Preliminaries}) which are applied in FLUX mdoel, next we propose our main contribution: Entropy-based Attention Weighting (\cref{Entropy}) and Agent framework (\cref{agent}).

\begin{figure*}[htb]
    \centering
    \includegraphics[width=\textwidth]{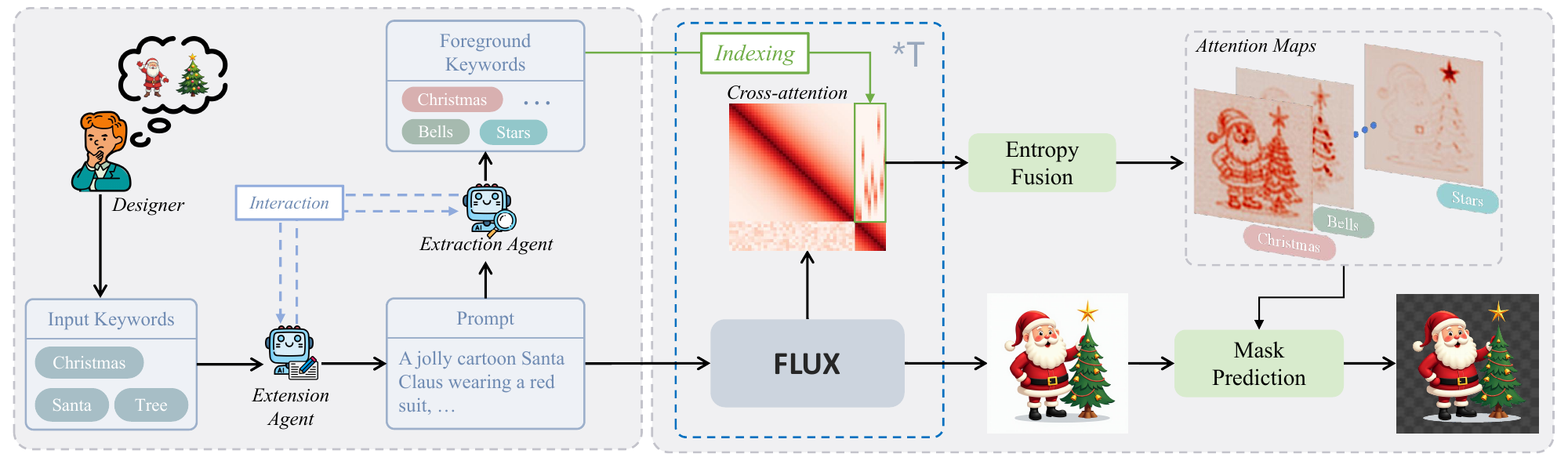}
    \caption{The overview of our pipeline includes agents responsible for prompt extension and keyword extraction. The prompt serves as the input for the pretrained FLUX model, while the extracted keywords guide the alpha channel estimation using an entropy-based feature fusion strategy, ultimately facilitating the RGBA image generation process.}
    \label{fig:model}
    \centering

\end{figure*}

\subsection{Overview}
\label{overview}
As illustrated in \cref{fig:model}, a fully automated framework is proposed for subject-centric generation:
(1) Our method receives elements input by the user (\textit{e.g., Christmas, Santa, Tree}), utilizes an agent based on LLMs to automatically expand these elements into an expressive prompt, and extracts nouns related to primary elements (\textbf{e.g., Christmas, Bells, Stars \etc}).
(2) The expanded prompt is used for the conditional generation of the pre-trained text-to-image model Flux, producing three-channel RGB images denoted as $\hat{I}\in \mathcal{R}^{H\times W\times 3}$, where $H$ and $W$ defines the height and width of the image $\hat{I}$. Concurrently, in each diffusion reverse step $t$, we use keywords to index corresponding parts of each cross-attention map, which are then integrated using our proposed entropy-based weighting strategy.
(3) We apply the merged cross-attention map as guidance to precisely predict the mask of the primary subject $M\in \mathcal{R}^{H\times W\times 1}$ that selectively removing irrelevant objects and preserving primary elements. %$I$ and $\alpha$ is concatenated, enabling precise RGBA image generation through this process.

\subsection{Diffusion Model}
\label{Preliminaries}
\textbf{Conditional Rectified Flow} Recent advancements in text-to-image generation are moving towards using rectified flow \cite{liu2022flow, albergo2022building} to construct an ordinary differential equation (ODE) that transports between two empirically observed distributions, $ x_0 $  and $  x_1  $ . In an ODE-based diffusion model, the evolution process is specified to follow the ODE: $\frac{dx}{dt} = f(x, t)$.
Here, the focus is on approximating $ \frac{dx}{dt} $ by a model $v_\theta(x, t)$ with trainable parameters $\theta$ , which represents the velocity at which $ x_0 $  evolves towards $ x_1  $.  
As a simple example, let us assume the evolution from $  x_0 $  to $  x_1 $  is a straight line, especially when $ x_1 $ follows the standard distribution, we have:
\begin{equation}
x_t = (1 - t) x_0 + t \, \epsilon, \quad \text{where} \quad \epsilon \sim \mathcal{N}(0, I).
\end{equation}
Moreover, \cite{lipman2022flow, rectified_flow} construct a conditional vector filled $u_t(x \mid \epsilon)$ which corresponds to a target probability density path $p_t (x \mid \epsilon)$. In such a way, it provides a objective known as Conditional Flow Matching (CFM): 
\begin{equation}
\mathcal{L}_{C F M}=\mathbb{E}_{t, p_t(x \mid \epsilon), p(\epsilon)}\left\|v_{\theta}(x, t)-u_t(x \mid \epsilon)\right\|_2^2.
\end{equation}
\textbf{MM-DiT} To construct $v_{\Theta}(x, t)$,  the previously popular combination of UNet and Transformer models commonly used in diffusion models is increasingly being replaced by the DiT (Diffusion Transformer) architecture \cite{dit}. Recently, numerous models have adopted this approach, achieving notable results \cite{Stability2024,BlackForest2024,sd35}.

In order to better align text information with image information, MM-DiT does not directly apply cross-attention between the extracted text features and the features of each image layer as in many previous methods \cite{latentdiffusion, podell2024sdxl}. Instead, it designs a dual-stream model: different trainable layers are assigned to different data modalities, and then they are concatenated for the attention operation. 
Recent studies ~\cite{Stability2024,BlackForest2024,sd35} has proven that the MM-DiT architecture is effective for text-to-image generation.
In our work, we utilize and explore the latest MM-DiT-based text-to-image pre-trained model, Flux~\cite{BlackForest2024}.

\subsection{Attention Extraction and Entropy-based Attention Weighing}
\label{Entropy}
Differ from previous pretrained text-to-image diffusion model \cite{latentdiffusion, podell2024sdxl, chen2024pixartalpha},
where cross-attention and self-attention are applied at different layers, Flux operates differently. It contains 19 dual-stream MM-DiT blocks and 38 single-stream DiT blocks. In dual-stream blocks, the image feature 
$z$ and text embeddings $e$ are processed separately, through different linear, scaling, and normalization layers. Afterward, they are concatenated and passed through the attention operation.
In single-stream blocks, the image feature 
$z$ and text embeddings are processed by the same transformers.

Since for each attention layer $l$, the image feature $z$ and the text feature $e$ are concatenated together for attention operation, the same strategy is applied consistently across all attention maps. Moreover, the attention map can be considered as a combination of the self-attention map and the cross-attention map. Similar to \cite{tang2022daam, quattrini2024alfie}, we specifically extract the cross-attention map to highlight the influence of key words.
 % e$ are concatenated first and then fed into the attention layer directly.
Formally, for a specific $t$ and $l$, the attention map is defined as $\mathcal{A}_{S}^{t,l} \in \mathbb{R}^{(hw + N) \times (hw + N) \times f}$, where $h$ and $w$ represent the latent spatial dimensions, $N$ represents the length of the text tokens, and $f$ defines the number of feature channels.
\begin{equation}
    \mathcal{A}_{S}^{t,l} := softmax(Q_{cat(z,e)}K_{cat(z,e)}/\sqrt{d}),
\end{equation}
where $\sqrt{d}$ stands for the scaling factor. Finally, the cross-attention $\mathcal{A}_{C}^{t,l} \in \mathbb{R}^{N \times hw \times C}$ can then extracted from $\mathcal{A}_{S}^{t,l}$ (see Fig.~\ref{fig:model}):
\begin{equation}
    \mathcal{A}_{C}^{t,l} := \mathcal{A}_{S}^{t,l}[N:, :N, :].
\end{equation}

\begin{figure}[!ht]
\begin{center}
\includegraphics[width=\linewidth]{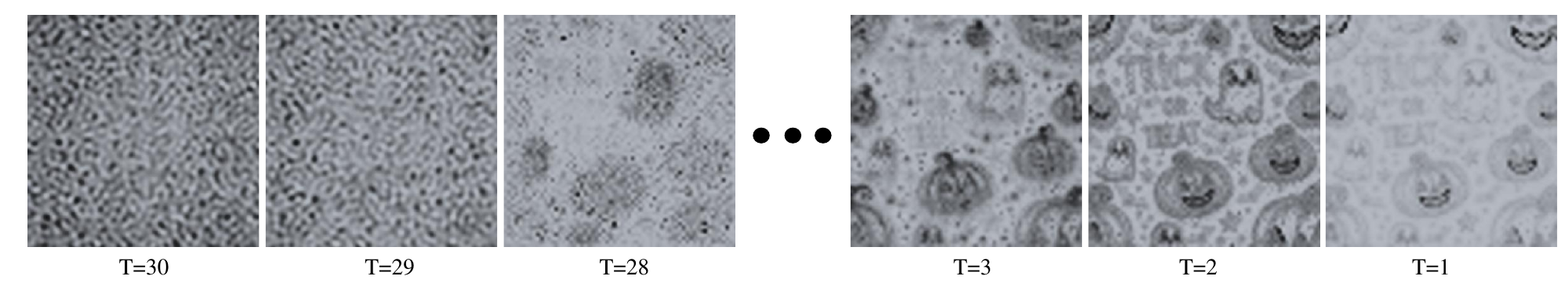}
\end{center}
\caption{Illustration of the attention map of the FLUX model for the generation process with the prompt "trick or treat" across different denoising steps. We use the Euler distance sampling method. The results show that at different timesteps, the noise contains varying amounts of information.}
\label{fig:figure_attention}
\end{figure}

 We observe that the information content in the cross-attention maps is not equal across different time steps and different layers. A simple accumulation might lead to suboptimal subject mask estimation. This is due to the unequal contribution of information from each time step and layer, potentially diluting more informative features(see Fig.\ref{fig:figure_attention}).
Thus, we proposed an entropy-based attention map weighting strategy, assessing the information content of attention maps at different time steps and layers, which allows to assign greater weights to more informative maps during the weighting process.
Specifically, for a given time $t$, layer $l$, and text token $n$'s attention map$\mathcal{A}_{C}^{t,l,n}$, we calculate its Histogram Probability $P_{t,l,n}$, 
 then the entropy $H_{t,l}$ of $\mathcal{A}_{C}^{t,l,n}$ can be obtain as follows:
\begin{equation}
    H_{t,l} = \frac{1}{N} \sum_{n=1}^{N} (-P_{t,l,n} \log_2 P_{t,l,n}).
\end{equation}
Ultimately, we can perform weighting on each attention map $\mathcal{A}_{C}^{t,l}$ across all $T$ steps and all $L$ layers:
\begin{equation}
    W_{\mathcal{A}_{C}^{t,l}} = \frac{1}{H_{t,l} + 1e-6} ,
\end{equation}
\begin{equation}
    \mathcal{A}_{C} =  \frac{1}{TL} \sum_{l=1}^{L} \sum_{t=1}^{T} W_{\mathcal{A}_{C}^{t,l}}\times \mathcal{A}_{C}^{t,l}.
\end{equation}
Through this method, we can obtain the cross-attention map $\mathcal{A}_{C}$ corresponding to prompt $e$. 
 We then select primary elements $\hat{n}_{fg} = [\hat{n}_{1}, ..., \hat{n}_{\hat{N}}]$ and extract their associated attention maps $\mathcal{A}_{C}^{\hat{n}_{fg}}$ from $\mathcal{A}_{C}$  to guide alpha estimation. Finally, these attention maps are organized into a 4-value map, which is then used in GrabCut\cite{rother2004grabcut} to guide the subject segmentation.
 As illustrated in \cref{fig:figure_attention}, we visualized the cross-attention maps produced by two fusion strategies: standard addition and entropy-weighting.  The resulting attention maps for each keyword and final generated images demonstrate that using an entropy-weighting strategy yields a clearer cross-attention map with reduced noise influence.

We observe that the highlighted regions in the feature map are strongly associated with the word, which impacts the accuracy of our decision on which object to retain. Words irrelevant to primary elements can affect the accuracy of alpha channel estimation. To optimize prompt composition and keyword extraction for primary elements, we introduce an agent framework to address this challenge.

\subsection{Agent-based prompt extension and keywords extraction}
\label{agent}
To enhance the effect of our approach on subject-centric generation, 
we propose an agent-driven framework, which can output higher quality prompt and elements through reflection \cite{shinn2024reflexion} and multi-agent collaboration. The framework extends key elements or casual user input \( K \) into a detailed scene prompt \( \hat{P} \) with enriched elements for image generation, then extracts primary elements \( \hat{n}_{fg} \) from the prompt to enable accurate extraction of the required attention map \( \mathcal{A}_{C}^{\hat{n}_{fg}} \). The whole framework is illustrated in Alg.~\ref{alg:agent}.%Specifically, the framework consists of the Extension and Extraction Agents, as illustrated in Alg.~\ref{alg:agent}.

\begin{algorithm}
\caption{The algorithm of multi-agent framework.}
\label{alg:agent}
\begin{algorithmic}[1]
\Require Keywords $K$
\Ensure Prompt $\hat{P}$ and Foreground Keywords $\hat{n}_{fg}$
\State Initialize Expander, Optimizer, Extractor, Filter: $M_{exp}$, $M_{opt}$, $M_{ext}$, $M_{flt}$
\Function{Extension\_Agent}{$K[, P]$}
    \If{input only have $K$}
        \State $P \gets M_{exp}(K)$
    \EndIf
    \State $\hat{P} \gets M_{opt}(K, P)$
    \While {$M_{opt}$ think need revision}
        \State $\hat{P} \gets M_{opt}(K, \hat{P})$
    \EndWhile
    \State \textbf{return} $\hat{P}$
\EndFunction
\Function{Extraction\_Agent}{$K, \hat{P}$}
    \State $n_{fg} \gets M_{ext}(\hat{P})$
    \State $\hat{n}_{fg} \gets M_{flt}(\hat{P}, n_{fg})$
    \While{$M_{ext}$ or $M_{flt}$ think need revision}
        \State $\hat{P} \gets \textsc{Extension\_Agent}(K, \hat{P})$
        \State $n_{fg} \gets M_{ext}(\hat{P})$
        \State $\hat{n}_{fg} \gets M_{flt}(\hat{P}, n_{fg})$
    \EndWhile
    \State \textbf{return} $\hat{P}, \hat{n}_{fg}$
\EndFunction

\State $\hat{P} \gets \textsc{Extension\_Agent}(K)$
\State $\hat{P}, \hat{n}_{fg} \gets \textsc{Extraction\_Agent}(K, \hat{P})$

\State \textbf{return} $\hat{P}, \hat{n}_{fg}$
\end{algorithmic}
\end{algorithm}

% \paragraph{Extension Agent.} 
In the Extension Agent, given input $K$, $M_{exp}$ expands $K$ into more expressive prompt $P$. To ensure the quality of prompt $P$, we introduce the self-reflection in prompt generation. The optimizer $M_{opt}$ continuously iteratively optimizes $P$ based on $K$ until the $M_{opt}$ determines that the optimized prompt is good enough and finally outputs $\hat{P}$.

After getting the optimized prompt $\hat{P}$, in the Extraction Agent, the extractor $M_{ext}$  checks $M_{opt}$'s output and extracts nouns $n_{fg} = [n_{1}, ..., n_{\hat{N}}]$. To filter out abstract terms that could misalign the attention map and affect alpha estimation, we apply a re-extract strategy: the filter $M_{flt}$ removes abstract words like \textit{``colorful'', ``details", ``data", etc}. 
and outputs filtered primary elements' vocabulary $\hat{n}_{fg} = [\hat{n}_{1}, ..., \hat{n}_{\hat{N}}]$. 
The Extraction Agent can send a revision request to the Extension Agent if $M_{ext}$ or $M_{flt}$ thinks that $\hat{P}$ or $n_{fg}$ has too many abstract elements, the Extension Agent is required to re-provide the prompt $\hat{P}$ to ensure the overall generation quality of the subject segmentation and the generated image.
%and filters if the $n_{fg}$ contains abstract words such as \textit{``colorful'', ``details", ``data", etc}. These words are harmful for alpha estimation with associated attention map between the image and words. 
% \begin{figure*}[htp]
% \begin{center}
% \includegraphics[width=\textwidth]{sec/figures/main_result.pdf}
% \end{center}
% \caption{Qualitative comparison of our method with two types of methods: Existing RGBA image generation and Matting Method.}
% \label{fig:figure2}
% \end{figure*}

\begin{figure*}[ht]
\centering
    \begin{subfigure}[b]{\linewidth}
        \centerline{\includegraphics[width=0.9\linewidth]{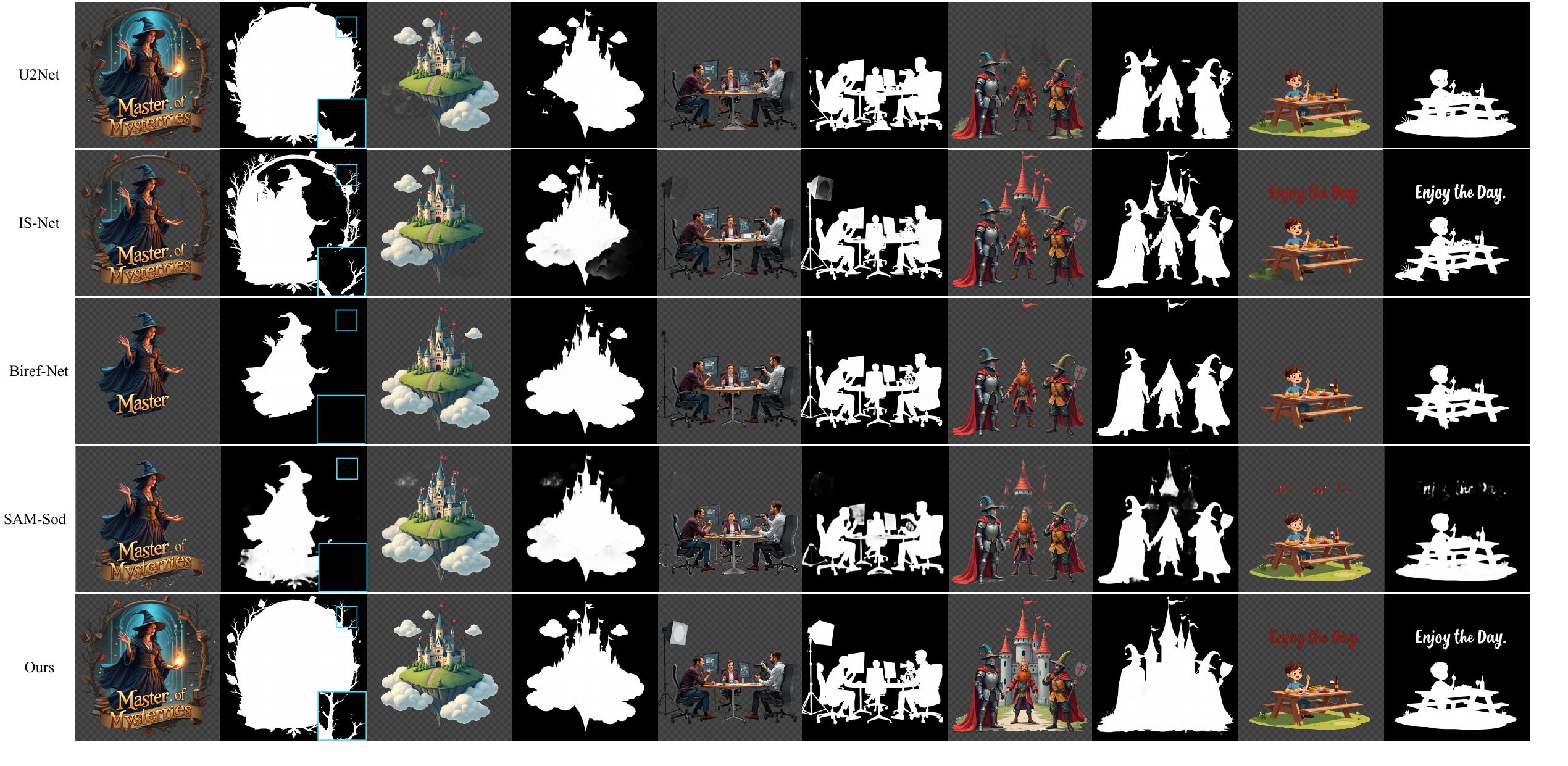}}
        \caption{Direct comparison with generation-then-segmentation methods.}
        \label{fig:figure2_1}
    \end{subfigure}
    \hfill
    \begin{subfigure}[b]{\linewidth}
        \centerline{\includegraphics[width=0.9\linewidth]{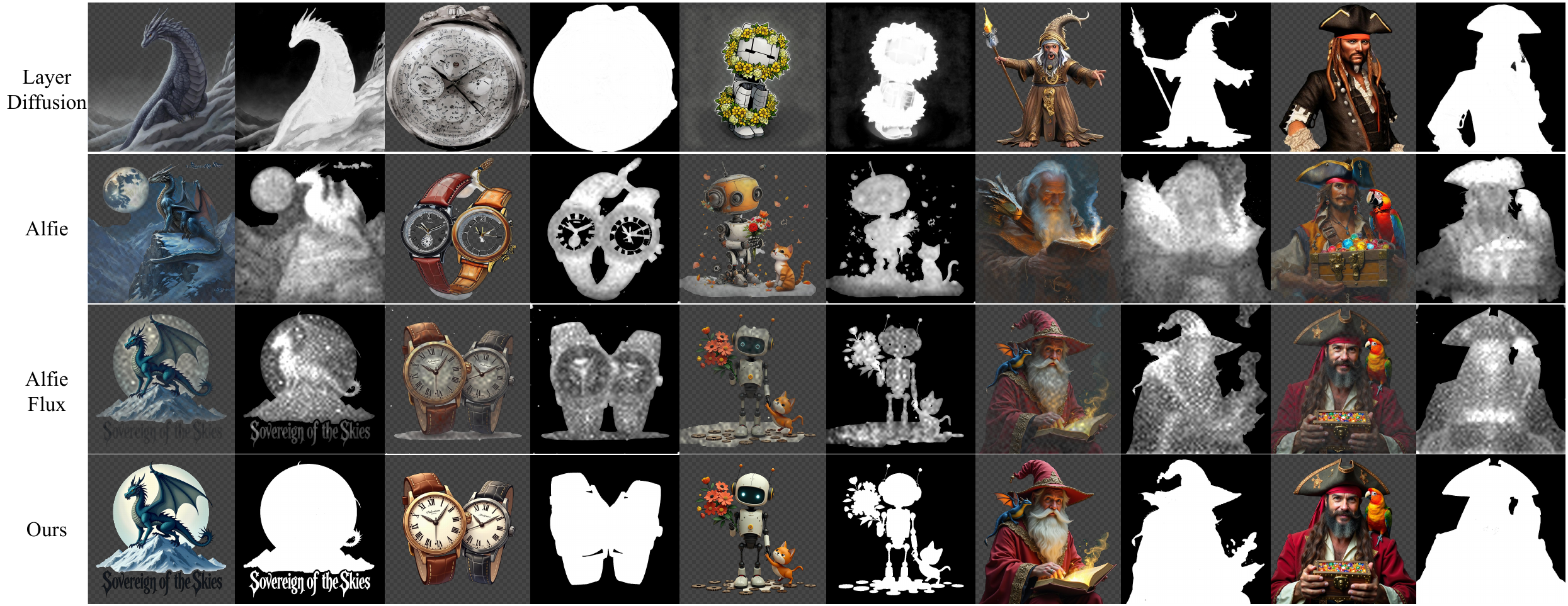}}
        \caption{Comparison with RGBA generation methods. Images are generated with the same prompt..}
        \label{fig:figure2_2}
    \end{subfigure}
\caption{Comparison with two pipelines: generation then segmentation/matting and RGBA image generation. Our method offers high-quality and precise generation of primary subject, outperforming other methods. }
\label{fig:figure2}
\end{figure*}

\section{Experiments}
\label{sec:experiments}
\subsection{Implementation Details} 
% 在实验中，我们使用了目前文生图的SOTA模型Flux，它包含57个attention processor layers,其中有19层为MM-DiT block，38层为single stream dit block。 因为这两种dit block输出的attention map结构相同，我们在feature extraction的流程中对他们进行了同样的处理我们在推理的过程中使用了EUlar Distance Sampling,并且去噪步数被设置为30 步。参考了Alfie的设定，我们使用0.8，0.2，0.1的阈值，将attention map转换为4-valuemap：sure foreground, probable foreground, probable background， and sure background，最终用Grabcut来预测alpha 通道。
In our experiments, we employed the current state-of-the-art text-to-image model, FLUX~\cite{BlackForest2024}.
%which includes 57 attention layers, with 19 layers being dual-stream MM-DiT blocks and 38 layers being single stream DiT blocks. 
As illustrated in \cref{Entropy}, these two types of DiT blocks produce attention maps in the same manner, we treated them identically during the feature extraction process. During inference, we applied Euler Distance Sampling, and the denoising steps were set to 30. Following ~\cite{quattrini2024alfie} setup, we used thresholds of 0.8, 0.2, and 0.1 to transform the attention map into a 4-value map: sure foreground, probable foreground, probable background, and sure background, ultimately using GrabCut to predict the mask of primary subject. In the agent framework, we use gpt-3.5-turbo\cite{ChatGPT} as the large language model. In gpt-3.5-turbo, temperature is 0.3, top p is 1, max tokens is 512.
% 我们对我们的方法做了detail的qualitative和quantatative的experiment。首先我们在 section（5.1）分别使用3组不带文字生成的prompt 和带文字生成的prompt进行了对比实验。 然后在section（5.2）中 我们进行了perceptual的user study和clip-score来quantatative的evaluate我们方法。最终在section（5.3）中，我们对设计的components进行了ablation study来展示这些模块的有效性

%In \cref{Quantitative}, we carried out a perceptual user study and used CLIP-scores \cite{clip} to quantitatively evaluate our method.	
%Subsequently, in \cref{Ablation}, we performed an ablation study on the designed components to demonstrate their effectiveness.
\begin{figure}[ht]
    \centering
    \includegraphics[width=\linewidth]{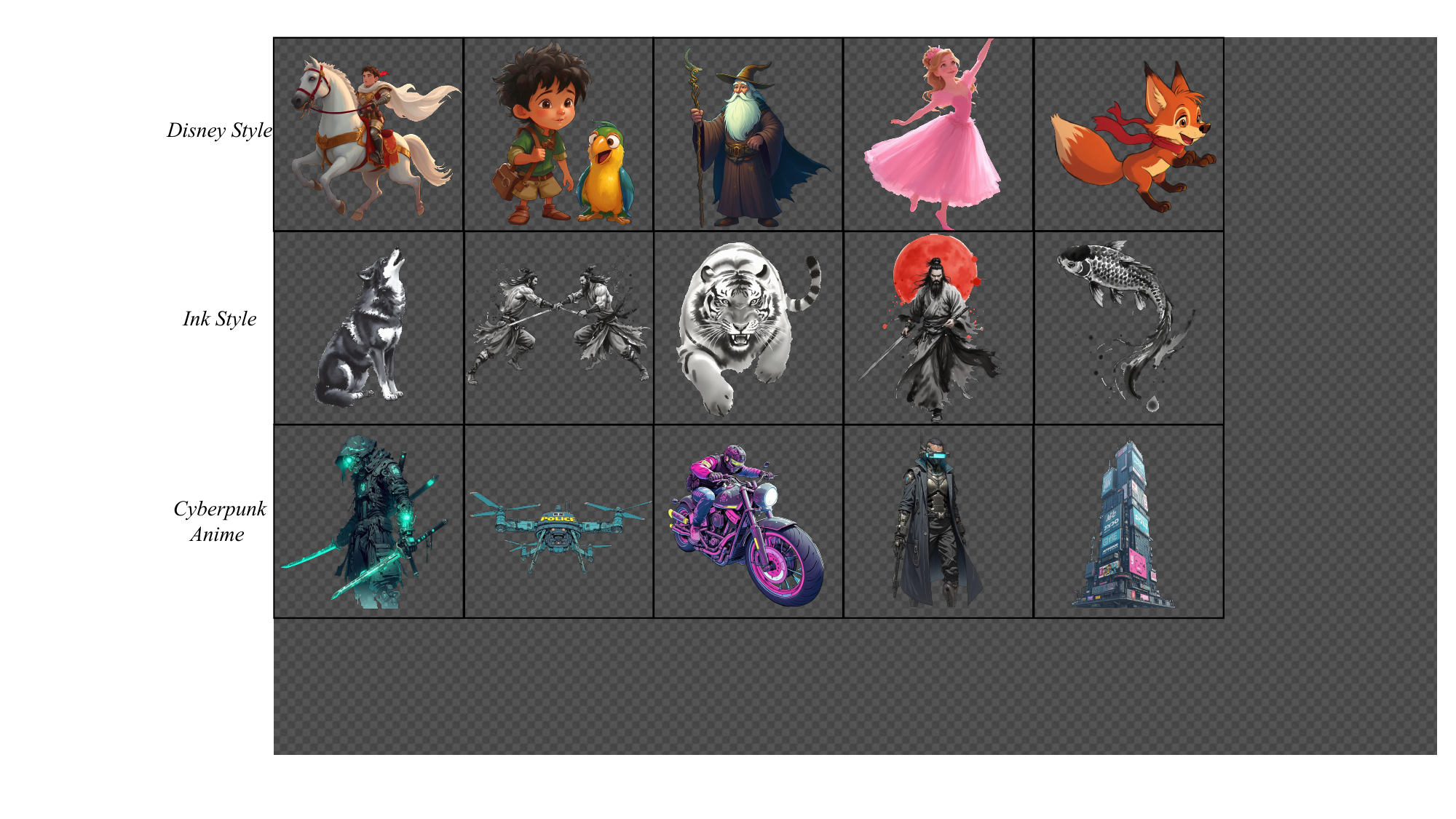}
    \caption{Results generated from our generation framework with different LoRA styles.}
    \label{fig:figurelora}
    \centering
\end{figure}

\subsection{Qualitative Results}
\label{Qualitative}

We conducted comprehensive qualitative and quantitative experiments to evaluate our method.
 We first performed comparative experiments using two different pipelines: generation-then segmentation/matting and direct RGBA generation. 
 
 For the first pipeline, we used FLUX to generate images and then applied state-of-the-art matting methods, including U2Net \cite{qin2020u2}, ISNet \cite{qin2022highly}, and BiRefNet \cite{zheng2024bilateral}. Additionally, we tested salient object detection methods like SAM-SOD \cite{gao2024multi}.
 It is worth mentioning that the pipeline most similar to our method follows a 'generation-then-salient-object-detection' approach. Both aim to segment the most prominent subject in an image.
Matting methods are trained to distinguish between the foreground and background, while segmentation methods focus on isolating the most salient object. However, both struggle to precisely delineate primary elements and retain complete information.
As shown in \cref{fig:figure2_1}, they erroneously remove essential text content due to the absence of text guidance, such as castle in the first column and cloud in the second column. 
In contrast, our approach, embedded in the flux attention layers, effectively captures high-response regions of key elements in text prompt during the generation process, thereby accurately identifying key elements and maximizing the preservation of complete information.

For the second pipeline, we compared our method with LayerDiffusion\cite{layerdiffusion} and Alfie\cite{quattrini2024alfie}. For a fair comparison, we also adapt Alfie in the Flux model, see Alfie-Flux in \cref{fig:figure2_2}.
It is important to note that we employed a zero-shot approach and did not train our model on a semi-transparent dataset. As a result, our method cannot handle semi-transparent objects in the same way as LayerDiffusion \cite{layerdiffusion}. 
In our experiments, for a fair comparison, the prompts are kept consistent across all methods.
According to the illustration in \cref{fig:figure2_2}, it is evident that Layer Diffusion, which necessitates comprehensive training procedures, encounters difficulties when adjusting to the most recent advancements in text-to-image model technologies. As a consequence, it generally produces images with only standard quality levels, without reaching superior outcomes. The Alfie and Alfie-Flux methods, while capable of making reasonably predictions of the alpha channel, exhibit limitations in their clarity. Additionally, these methods do not accurately capture details of primary elements, affecting image clarity. However, our method can generate images that align with the provided text and retain semantically relevant details.

To validate the diversity of our generation framework. we also test the Flux model with three LoRAs\cite{hu2022lora} : dsney style\footnote{Disney lora link: \url{https://huggingface.co/XLabs-AI/flux-lora-collection}}, ink style\footnote{Ink lora link: \url{https://civitai.com/models/73305/zyd232s-ink-style}} and cyperpunk anime style\footnote{Cyberpunk lora link: \url{https://civitai.com/models/128568/cyberpunk-anime-style}}. The results are shown in \cref{fig:figurelora} Our method consistently exhibited robust performance and successfully achieved the desired effects across all LoRA integrations, demonstrating its adaptability and effectiveness.

\subsection{Quantitative Results}
\label{Quantitative}
\begin{figure}[htb]
	\centering
		\includegraphics[width=\linewidth]{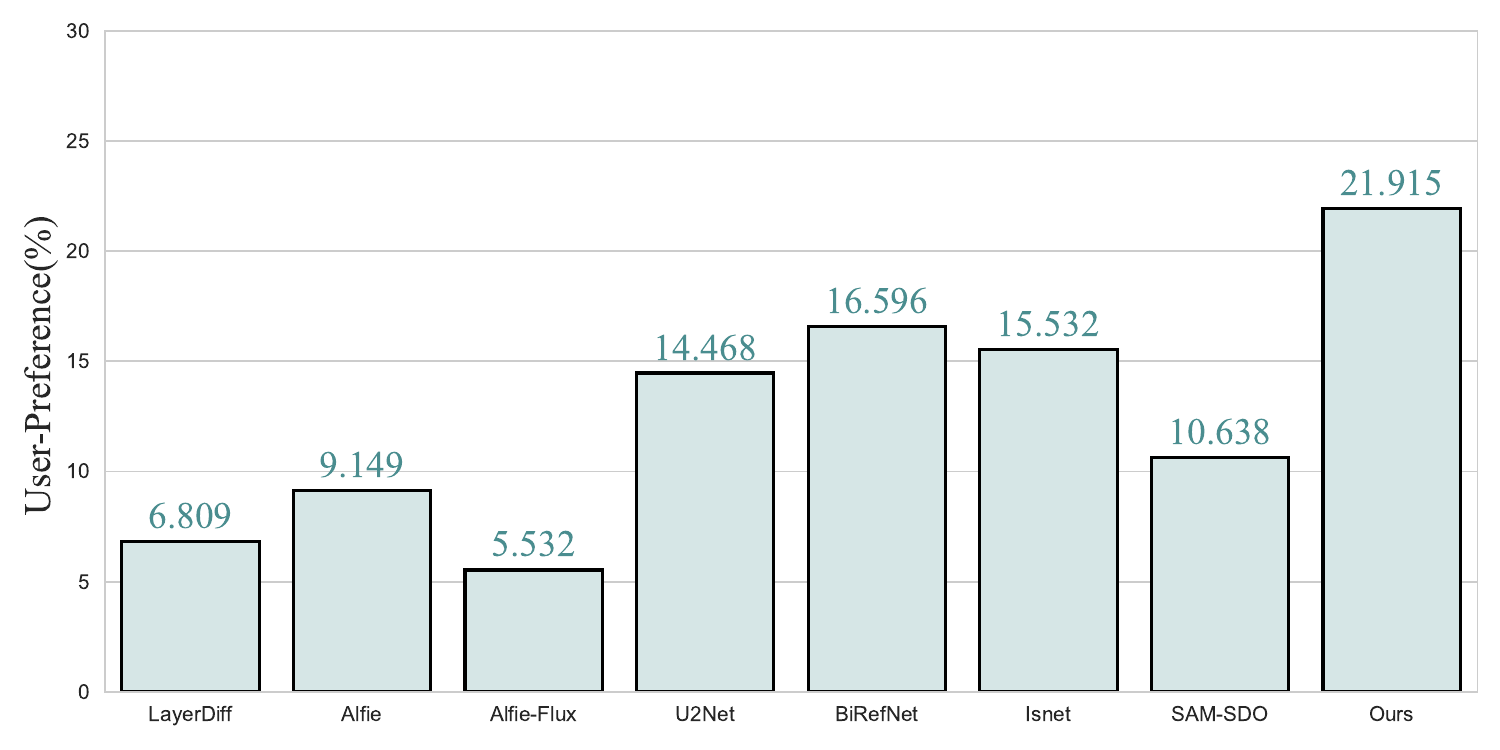}
            \caption{User preference. We compare RGBA image generation methods using the same prompts and evaluate matting methods using the same generated images.}
		\label{fig_E2_1}

	\label{fig_E2}
\end{figure}
\begin{table}[tb]\centering
    \caption{Clip-Score between generated primary elements and prompt for 4 image generation methods.}
    \label{tab:table_clip}
    \resizebox{0.48\textwidth}{!}{
    \large
    \begin{tabular}{*{5}{c}}
        \toprule
       Score & Layer-Diff &  Alfie & Alfie-Flux &  Ours  \\
        \midrule
        CLIP-Score & 24.92 & 30.78 & 30.88 & 31.73\\
        \bottomrule
    \end{tabular}
    }
\end{table}

To quantitatively evaluate our method, we first calculate the CLIP-Score (see \cref{tab:table_clip}). The results demonstrate that our approach achieves superior text-image alignment compared to other generation methods.
Additionally, we conducted a user study in which we randomly generated 20 sets of images and recruited 30 participants. Each participant was asked to compare our method with all possible pipelines and rank the generated images from best to worst based on their visual appeal and effectiveness. This study provided a direct assessment of the overall quality of the images produced by each approach. In our scoring system, the highest-ranked method received 8 points, while the lowest-ranked received 1 point. We then summed the scores for each method and normalized them by dividing by the total score, yielding the final user preference distribution.

As illustrated in \cref{fig_E2_1}, our method outperforms other RGBA generation approaches. This advantage can be attributed not only to its subject-centric generation capability but also to the strong generation ability of FLUX. Furthermore, when compared to the generation-then-matting pipeline using the same generated image, our method still achieves superior results. This is because it retains knowledge of the generation process, allowing it to better preserve the integrity of the main subject in the image.
%``Please rate the images from 1 to 9, the higher the score the tighter the connection is.'', which, along with the CLIP-Score, assessed the correlation between the generated images and the prompts. The second question, ``Which of the four RGBA images do you prefer?'' served to directly compare the visual appeal and effectiveness of the RGBA images generated by each method.
% 如图6所示， 我们展示了我们方法和3种生成方法对比的量化实验的结果。 图6 a 展示的是用户们对不同方法生成的RGBA图片和文字prompt之间的相关度，如图所示，我们的方法在满分为4分的情况下获得了xxx分，超出了第二名Alfie-Flux XX分。同时，第二个图展示了四种方法，生成图片和prompt之间计算的CLIP-SCORE（越大越好），我们的方法高于其他三种方法。这两个实验结果共同证明了我们方法生成图片非常adhere to text prompt。而对于另外一个user-study。我们发现，80%的用户都更喜欢我们的图片，展示出我们的方法略优于目前存在的RGBA图像生成sota方法

%As illustrated in \cref{fig_E2}, we present the results of quantitative experiments. \cref{fig_E2} displays the human perceptual correlation scores between the RGBA images generated by different methods and the text prompts, with our method scoring 3.69 out of a possible 4 points, surpassing the second method, Alfie-Flux, by 0.51 points. And \cref{tab:table_clip} shows the CLIP-Scores (higher is better) for images generated by the four methods in relation to the prompts, with our approach outperforming the other three, achieving score of 31.73. These results collectively demonstrate our method's superior adherence to text prompts. In another user study, we found that 67.46\% of participants preferred our images, indicating the advantage of our method over the current state-of-the-art RGBA image generation methods.

\begin{figure}[!t]
    \centering
    \includegraphics[width=\linewidth]{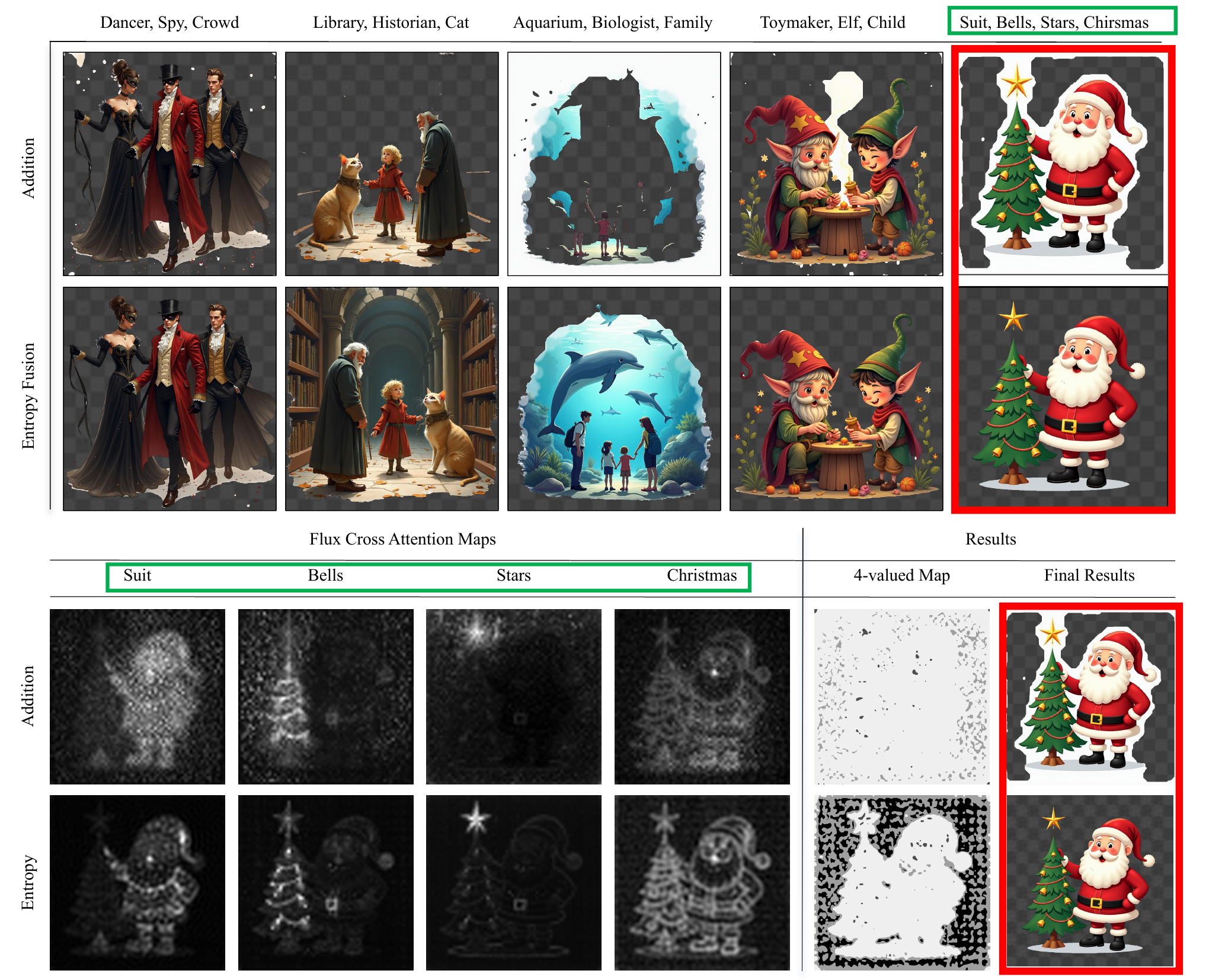}
    \caption{Comparison between two different feature fusion strategies: Addition and Entropy Fusion. The first image illustrates the final results of these two fusion methods, while the second image presents the cross-attention map corresponding to each word. Both experiments demonstrate that our entropy-based fusion method significantly enhances the generation of primary subjects.
    }
    \label{fig:figureablation}
    \centering
\end{figure}

\subsection{Ablation Study}
\label{Ablation}
\textbf{Entropy Based Feature Fusion.}
% 如图 7所示，我们展示了我们方法用两种不同attention fusion方法生成的结果。我们可以看到，使用addition将A融合成B的的策略可以提供粗略的前景alpha通道预测，会出现保留部分背景（第4列中部分白色区域被保留）或者将期望的前景去除的情况（第二列中“Library”对应的前景被去除）。而使用entropy-based的方法，由于能提供更精细的attention map，能精准的保留前景内容
%We first analyse the attention map of entropy based feature fusion and stard addition method, see \cref{fig:figure_attention}. 
As shown in ~\cref{fig:figureablation}, we demonstrate the results generated using two different attention fusion methods: the entropy-based fusion and the standard addition fusion. We first present the final generation results obtained using two methods (see the first figure in ~\cref{fig:figureablation}). Then, we illustrate the attention map throughout the generation process of the last column image (see the second figure in ~\cref{fig:figureablation}). The same image is highlighted with a red rectangle, while the same prompt is indicated with a green rectangle.

The addition-based fusion strategy for blending $\mathcal{A}_{C}^{t,l}$ into $\mathcal{A}_{C}$ provides a coarse generation of primary elements. This approach ignores the noise-to-information ratio of attention maps across different steps, leading to incomplete background removal (partial white areas in the first column) and the unintended removal of expected foreground elements (removal of the 'Library' foreground in the fourth column). The entropy-based fusion method addresses this issue by assigning higher weights to attention maps with greater information content, ensuring the accurate generation of foreground content without unexpected removal or retention.

\begin{figure}[htb]
\begin{center}
    \begin{subfigure}[b]{0.49\linewidth}
        \includegraphics[width=\linewidth]{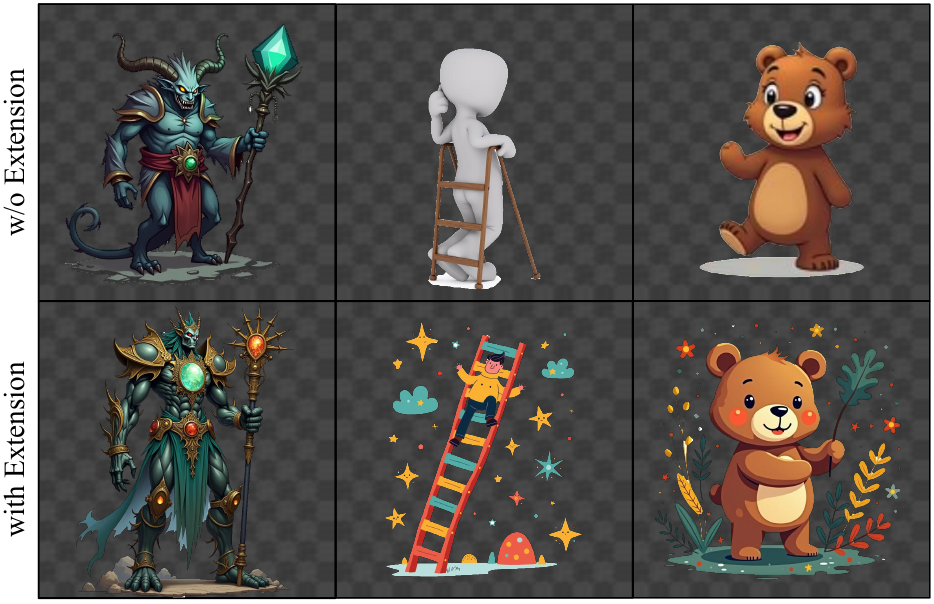}
        \caption{Prompt extension.}
        \label{fig:figure_prompt1}
    \end{subfigure}
    \hfill
    \begin{subfigure}[b]{0.49\linewidth}
        \includegraphics[width=\linewidth]{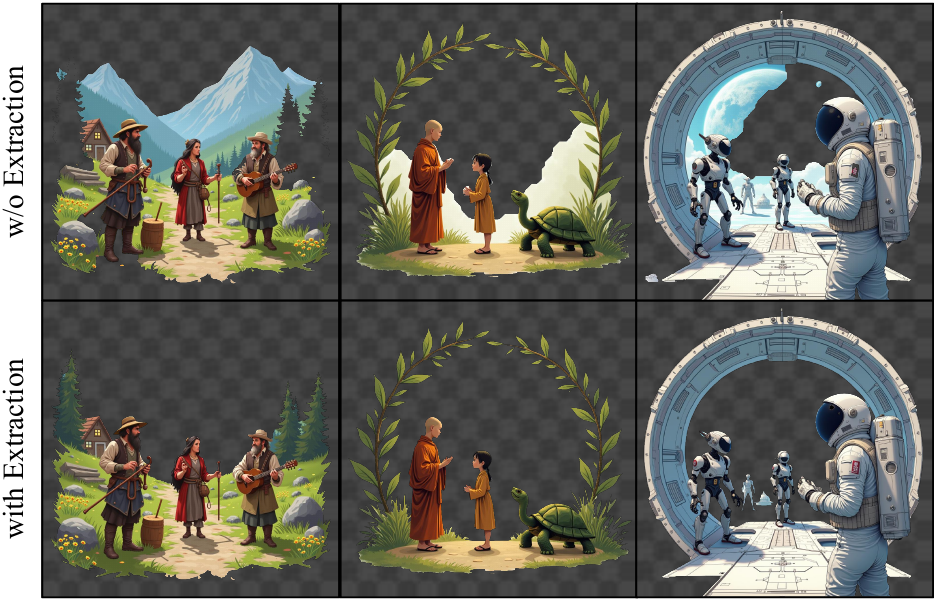}
        \caption{Keywords extraction.}
        \label{fig:figure_prompt2}
    \end{subfigure}
\end{center}
\caption{This figure illustrates the effects of prompt extension (a) on RGB image generation and the impact of keyword extraction (b) on mask prediction of primiary subjects.}
\label{fig:combined_figures}
\end{figure}

\textbf{Agent-based prompt extension and keyword extraction.}
In ~\cref{fig:combined_figures}, we first demonstrate the effects of generating images with and without the use of an extension. Specifically, for a set of elements, we uses these elements directly as prompts to generate images when no extension is applied. As shown in~\cref{fig:figure_prompt1}, images produced in this manner typically lack detail or are blurry. Conversely, when an extension is applied, the Extension agent enhances the descriptions of the elements, resulting in images with richer details.
In the ~\cref{fig:figure_prompt2}, we illustrate the impact of using an Extraction agent to identify foreground vocabulary on subject-centric image generation. Specifically, for a given prompt, without extraction, we eliminate some generic nouns for attention indexing. In contrast, with extraction, the Extraction agent identifies foreground-related vocabulary for attention indexing. As demonstrated in~\cref{fig:figure_prompt2}, without the use of an extraction agent, the generation of primary subject is suboptimal influenced by generic nouns not listed in the exclusion list. With the use of an extraction agent, we achieve more plausible generation of primary elements.
\noindent  
% \begin{figure}[htb]
% \begin{center}
%     \begin{subfigure}[b]{\linewidth}
%         \includegraphics[width=\linewidth]{sec/figures/prompt1.pdf}
%         \caption{prompt extension.}
%         \label{fig:figure_prompt1}
%     \end{subfigure}
%     \hfill
%     \begin{subfigure}[b]{\linewidth}
%         \includegraphics[width=\linewidth]{sec/figures/prompt2.pdf}
%         \caption{Keywords Extraction}
%         \label{fig:figure_prompt2}
%     \end{subfigure}
% \end{center}
% \caption{In this figure, we illustrate the effects of prompt extension (a) on RGB image generation and the impact of keyword extraction (b) on alpha channel prediction, respectively.}
% % 在子图的a和b中，我们分别展示了Prompt extension对RBG图片生成的影响和Keywords Extraction对Alpha通道预测的影响。
% \label{fig:combined_figures}
% \end{figure}

%%%%%%%%%%%%%%%%%%%%%%%%%%%%%%%%%%%
% ablation
% \input{}
% 5
%  Prompt Agent            |
%  不同feature block1
% block 2                  |
%   FFT                    |
%   Entropy Weighting      |
%  Final Results           |

%%%%%%%%%%%%%%%%%%%%%%%%%%%%%%%%%%%

%%%%%%%%%%%%%%%%%%%%%%%%%%%%%%%%%%%
% \input{c}
%%%%%%%%%%%%%%%%%%%%%%%%%%%%%%%%%%%
\section{ Conclusion}
\label{sec:conclusion}

%Our method follows a generation-then-segmentation paradigm, making it well-suited for subject-level customization tasks such as logos, stickers, emojis, and prints. However, compared to RGBA-based generation, our approach struggles with semi-transparent objects like smoke, water, and glass, as it lacks the ability to directly model transparency. A potential solution is to fine-tune the model using existing RGBA datasets, allowing it to better capture transparency variations and extend its applicability to a broader range of subject-centric generation tasks.

We propose a zero-shot generation pipeline capable of producing high-quality subject-centric images that seamlessly integrate with other visual content without requiring additional training. Our approach is built on two key components: an agent framework and an entropy-based attention map fusion module.
Extensive qualitative and quantitative experiments validate the effectiveness of our method and the functionality of its design modules. In summary, this work introduces a novel approach to generating high-quality subject-centric content, ensuring that only the desired elements are produced while eliminating unwanted content.

% \section{}
%In this work, we have explored the challenge of generating transparent images needed for visual content creation. 
%We designed a zero-shot generation pipeline, ZEBRA, capable of producing high-quality RGBA images that seamlessly integrate with other visual content without the need for additional training. In this research, we focus on two main components: an agent framework and an entropy-based attention map fusion module. These are crucial for user-friendly accurate primary elements generation with only simply inputs.
%Extensive qualitative and quantitative experiments validate the effectiveness of our method and the functionality of our design modules. In summary, this work introduces a novel approach to address a problem of generating transparent images that that is both highly important and practical, yet has previously received limited attention.
%We hope our work can broaden the understanding in this area and inspire continued exploration among researchers.

{
    \small
    \bibliographystyle{plainnat}
    \bibliography{arxiv}

\begin{thebibliography}{55}
\providecommand{\natexlab}[1]{#1}
\providecommand{\url}[1]{\texttt{#1}}
\expandafter\ifx\csname urlstyle\endcsname\relax
  \providecommand{\doi}[1]{doi: #1}\else
  \providecommand{\doi}{doi: \begingroup \urlstyle{rm}\Url}\fi

\bibitem[AI(2024{\natexlab{a}})]{Stability2024}
Stability AI.
\newblock stable-diffusion-3-medium.
\newblock \url{https://huggingface.co/stabilityai/stable-diffusion-3-medium}, 2024{\natexlab{a}}.
\newblock Accessed: 2024-11-03.

\bibitem[AI(2024{\natexlab{b}})]{sd35}
Stability AI.
\newblock stabilityai/stable-diffusion-3.5-large.
\newblock \url{https://huggingface.co/stabilityai/stable-diffusion-3.5-large}, 2024{\natexlab{b}}.
\newblock Accessed: 2024-11-03.

\bibitem[Albergo and Vanden-Eijnden(2022)]{albergo2022building}
Michael~S Albergo and Eric Vanden-Eijnden.
\newblock Building normalizing flows with stochastic interpolants.
\newblock \emph{arXiv preprint arXiv:2209.15571}, 2022.

\bibitem[Borji et~al.(2019)Borji, Cheng, Hou, Jiang, and Li]{borji2019salient}
Ali Borji, Ming-Ming Cheng, Qibin Hou, Huaizu Jiang, and Jia Li.
\newblock Salient object detection: A survey.
\newblock \emph{Computational visual media}, 5:\penalty0 117--150, 2019.

\bibitem[Brade et~al.(2023)Brade, Wang, Sousa, Oore, and Grossman]{brade2023promptify}
Stephen Brade, Bryan Wang, Mauricio Sousa, Sageev Oore, and Tovi Grossman.
\newblock Promptify: Text-to-image generation through interactive prompt exploration with large language models.
\newblock In \emph{Proceedings of the 36th Annual ACM Symposium on User Interface Software and Technology}, pages 1--14, 2023.

\bibitem[Cao et~al.(2023)Cao, Wang, Qi, Shan, Qie, and Zheng]{cao2023masactrl}
Mingdeng Cao, Xintao Wang, Zhongang Qi, Ying Shan, Xiaohu Qie, and Yinqiang Zheng.
\newblock Masactrl: Tuning-free mutual self-attention control for consistent image synthesis and editing.
\newblock In \emph{Proceedings of the IEEE/CVF International Conference on Computer Vision}, pages 22560--22570, 2023.

\bibitem[Chen et~al.(2024{\natexlab{a}})Chen, YU, GE, Yao, Xie, Wang, Kwok, Luo, Lu, and Li]{chen2024pixartalpha}
Junsong Chen, Jincheng YU, Chongjian GE, Lewei Yao, Enze Xie, Zhongdao Wang, James Kwok, Ping Luo, Huchuan Lu, and Zhenguo Li.
\newblock Pixart-\${\textbackslash}alpha\$: Fast training of diffusion transformer for photorealistic text-to-image synthesis.
\newblock In \emph{The Twelfth International Conference on Learning Representations}, 2024{\natexlab{a}}.

\bibitem[Chen et~al.(2024{\natexlab{b}})Chen, Huang, Liu, Shen, Zhao, and Zhao]{chen2024anydoor}
Xi Chen, Lianghua Huang, Yu Liu, Yujun Shen, Deli Zhao, and Hengshuang Zhao.
\newblock Anydoor: Zero-shot object-level image customization.
\newblock In \emph{Proceedings of the IEEE/CVF conference on computer vision and pattern recognition}, pages 6593--6602, 2024{\natexlab{b}}.

\bibitem[Couairon et~al.(2023)Couairon, Verbeek, Schwenk, and Cord]{couairon2023diffedit}
Guillaume Couairon, Jakob Verbeek, Holger Schwenk, and Matthieu Cord.
\newblock Diffedit: Diffusion-based semantic image editing with mask guidance.
\newblock In \emph{The Eleventh International Conference on Learning Representations}, 2023.

\bibitem[Crowson et~al.(2024)Crowson, Baumann, Birch, Abraham, Kaplan, and Shippole]{crowson2024scalable}
Katherine Crowson, Stefan~Andreas Baumann, Alex Birch, Tanishq~Mathew Abraham, Daniel~Z Kaplan, and Enrico Shippole.
\newblock Scalable high-resolution pixel-space image synthesis with hourglass diffusion transformers.
\newblock In \emph{Forty-first International Conference on Machine Learning}, 2024.

\bibitem[Esser et~al.()Esser, Kulal, Blattmann, Entezari, M{\"u}ller, Saini, Levi, Lorenz, Sauer, Boesel, et~al.]{rectified_flow}
Patrick Esser, Sumith Kulal, Andreas Blattmann, Rahim Entezari, Jonas M{\"u}ller, Harry Saini, Yam Levi, Dominik Lorenz, Axel Sauer, Frederic Boesel, et~al.
\newblock Scaling rectified flow transformers for high-resolution image synthesis, march 2024.
\newblock \emph{URL http://arxiv. org/abs/2403.03206}.

\bibitem[Gao et~al.(2024)Gao, Zhang, Yan, and Lu]{gao2024multi}
Shixuan Gao, Pingping Zhang, Tianyu Yan, and Huchuan Lu.
\newblock Multi-scale and detail-enhanced segment anything model for salient object detection.
\newblock In \emph{Proceedings of the 32nd ACM International Conference on Multimedia}, pages 9894--9903, 2024.

\bibitem[Hertz et~al.(2023{\natexlab{a}})Hertz, Mokady, Tenenbaum, Aberman, Pritch, and Cohen-or]{hertz2023prompt}
Amir Hertz, Ron Mokady, Jay Tenenbaum, Kfir Aberman, Yael Pritch, and Daniel Cohen-or.
\newblock Prompt-to-prompt image editing with cross-attention control.
\newblock In \emph{The Eleventh International Conference on Learning Representations}, 2023{\natexlab{a}}.

\bibitem[Hertz et~al.(2023{\natexlab{b}})Hertz, Mokady, Tenenbaum, Aberman, Pritch, and Cohen-or]{hertz2023prompttoprompt}
Amir Hertz, Ron Mokady, Jay Tenenbaum, Kfir Aberman, Yael Pritch, and Daniel Cohen-or.
\newblock Prompt-to-prompt image editing with cross-attention control.
\newblock In \emph{The Eleventh International Conference on Learning Representations}, 2023{\natexlab{b}}.

\bibitem[Ho et~al.(2020)Ho, Jain, and Abbeel]{ho2020denoising}
Jonathan Ho, Ajay Jain, and Pieter Abbeel.
\newblock Denoising diffusion probabilistic models.
\newblock In \emph{Advances in neural information processing systems}, pages 6840--6851, 2020.

\bibitem[Hu et~al.(2022)Hu, Shen, Wallis, Allen-Zhu, Li, Wang, Wang, Chen, et~al.]{hu2022lora}
Edward~J Hu, Yelong Shen, Phillip Wallis, Zeyuan Allen-Zhu, Yuanzhi Li, Shean Wang, Lu Wang, Weizhu Chen, et~al.
\newblock Lora: Low-rank adaptation of large language models.
\newblock \emph{ICLR}, 1\penalty0 (2):\penalty0 3, 2022.

\bibitem[Huang et~al.(2024)Huang, Wen, Zhao, Hu, Liu, Jia, Mao, Wang, Zhang, Chen, et~al.]{huang2024subjectdrive}
Binyuan Huang, Yuqing Wen, Yucheng Zhao, Yaosi Hu, Yingfei Liu, Fan Jia, Weixin Mao, Tiancai Wang, Chi Zhang, Chang~Wen Chen, et~al.
\newblock Subjectdrive: Scaling generative data in autonomous driving via subject control.
\newblock \emph{arXiv preprint arXiv:2403.19438}, 2024.

\bibitem[Kirillov et~al.(2023)Kirillov, Mintun, Ravi, Mao, Rolland, Gustafson, Xiao, Whitehead, Berg, Lo, et~al.]{kirillov2023segment}
Alexander Kirillov, Eric Mintun, Nikhila Ravi, Hanzi Mao, Chloe Rolland, Laura Gustafson, Tete Xiao, Spencer Whitehead, Alexander~C Berg, Wan-Yen Lo, et~al.
\newblock Segment anything.
\newblock In \emph{Proceedings of the IEEE/CVF International Conference on Computer Vision}, pages 4015--4026, 2023.

\bibitem[Labs(2024)]{BlackForest2024}
Black~Forest Labs.
\newblock Flux.1 [dev] model card, flux.1-dev.
\newblock \url{https://huggingface.co/black-forest-labs/FLUX.1-dev}, 2024.
\newblock Accessed: 2024-11-03.

\bibitem[Li et~al.(2023)Li, Huang, Ding, and Li]{li2023layerdiffusion}
Pengzhi Li, Qinxuan Huang, Yikang Ding, and Zhiheng Li.
\newblock Layerdiffusion: Layered controlled image editing with diffusion models.
\newblock In \emph{SIGGRAPH Asia 2023 Technical Communications}, pages 1--4. 2023.

\bibitem[Lipman et~al.(2022)Lipman, Chen, Ben-Hamu, Nickel, and Le]{lipman2022flow}
Yaron Lipman, Ricky~TQ Chen, Heli Ben-Hamu, Maximilian Nickel, and Matt Le.
\newblock Flow matching for generative modeling.
\newblock \emph{arXiv preprint arXiv:2210.02747}, 2022.

\bibitem[Liu et~al.(2024)Liu, Wang, Cao, Jia, and Huang]{liu2024towards}
Bingyan Liu, Chengyu Wang, Tingfeng Cao, Kui Jia, and Jun Huang.
\newblock Towards understanding cross and self-attention in stable diffusion for text-guided image editing.
\newblock In \emph{Proceedings of the IEEE/CVF Conference on Computer Vision and Pattern Recognition}, pages 7817--7826, 2024.

\bibitem[Liu and Chilton(2022)]{liu2022design}
Vivian Liu and Lydia~B Chilton.
\newblock Design guidelines for prompt engineering text-to-image generative models.
\newblock In \emph{Proceedings of the 2022 CHI conference on human factors in computing systems}, pages 1--23, 2022.

\bibitem[Liu et~al.(2023{\natexlab{a}})Liu, Vermeulen, Fitzmaurice, and Matejka]{liu20233dall}
Vivian Liu, Jo Vermeulen, George Fitzmaurice, and Justin Matejka.
\newblock 3dall-e: Integrating text-to-image ai in 3d design workflows.
\newblock In \emph{Proceedings of the 2023 ACM designing interactive systems conference}, pages 1955--1977, 2023{\natexlab{a}}.

\bibitem[Liu et~al.(2022)Liu, Gong, and Liu]{liu2022flow}
Xingchao Liu, Chengyue Gong, and Qiang Liu.
\newblock Flow straight and fast: Learning to generate and transfer data with rectified flow.
\newblock \emph{arXiv preprint arXiv:2209.03003}, 2022.

\bibitem[Liu et~al.(2023{\natexlab{b}})Liu, Zhang, Shen, Zheng, Zhu, Feng, Liu, Zhao, Zhou, and Cao]{liu2023customizable}
Zhiheng Liu, Yifei Zhang, Yujun Shen, Kecheng Zheng, Kai Zhu, Ruili Feng, Yu Liu, Deli Zhao, Jingren Zhou, and Yang Cao.
\newblock Customizable image synthesis with multiple subjects.
\newblock \emph{Advances in neural information processing systems}, 36:\penalty0 57500--57519, 2023{\natexlab{b}}.

\bibitem[Mou et~al.(2024)Mou, Wang, Xie, Wu, Zhang, Qi, and Shan]{mou2024t2i}
Chong Mou, Xintao Wang, Liangbin Xie, Yanze Wu, Jian Zhang, Zhongang Qi, and Ying Shan.
\newblock T2i-adapter: Learning adapters to dig out more controllable ability for text-to-image diffusion models.
\newblock In \emph{Proceedings of the AAAI Conference on Artificial Intelligence}, pages 4296--4304, 2024.

\bibitem[Nichol and Dhariwal(2021)]{nichol2021improved}
Alexander~Quinn Nichol and Prafulla Dhariwal.
\newblock Improved denoising diffusion probabilistic models.
\newblock In \emph{International conference on machine learning}, pages 8162--8171, 2021.

\bibitem[OpenAI(2023)]{ChatGPT}
OpenAI.
\newblock Chatgpt 3.5.
\newblock \url{https://chatgpt.com/}, 2023.
\newblock Accessed: 2024-11-03.

\bibitem[Peebles and Xie(2023{\natexlab{a}})]{dit}
William Peebles and Saining Xie.
\newblock Scalable diffusion models with transformers, 2023{\natexlab{a}}.

\bibitem[Peebles and Xie(2023{\natexlab{b}})]{peebles2023scalable}
William Peebles and Saining Xie.
\newblock Scalable diffusion models with transformers.
\newblock In \emph{Proceedings of the IEEE/CVF International Conference on Computer Vision}, pages 4195--4205, 2023{\natexlab{b}}.

\bibitem[PIKA(2024)]{pika}
PIKA.
\newblock Pika.
\newblock \url{https://pika.art/login}, 2024.
\newblock Accessed: 2025-02-27.

\bibitem[Podell et~al.(2024)Podell, English, Lacey, Blattmann, Dockhorn, M{\"u}ller, Penna, and Rombach]{podell2024sdxl}
Dustin Podell, Zion English, Kyle Lacey, Andreas Blattmann, Tim Dockhorn, Jonas M{\"u}ller, Joe Penna, and Robin Rombach.
\newblock Sdxl: Improving latent diffusion models for high-resolution image synthesis.
\newblock In \emph{The Twelfth International Conference on Learning Representations}, 2024.

\bibitem[Qin et~al.(2020)Qin, Zhang, Huang, Dehghan, Zaiane, and Jagersand]{qin2020u2}
Xuebin Qin, Zichen Zhang, Chenyang Huang, Masood Dehghan, Osmar~R Zaiane, and Martin Jagersand.
\newblock U2-net: Going deeper with nested u-structure for salient object detection.
\newblock \emph{Pattern recognition}, 106:\penalty0 107404, 2020.

\bibitem[Qin et~al.(2022{\natexlab{a}})Qin, Dai, Hu, Fan, Shao, and Van~Gool]{isnet}
Xuebin Qin, Hang Dai, Xiaobin Hu, Deng-Ping Fan, Ling Shao, and Luc Van~Gool.
\newblock Highly accurate dichotomous image segmentation.
\newblock In \emph{European Conference on Computer Vision}, pages 38--56. Springer, 2022{\natexlab{a}}.

\bibitem[Qin et~al.(2022{\natexlab{b}})Qin, Dai, Hu, Fan, Shao, and Van~Gool]{qin2022highly}
Xuebin Qin, Hang Dai, Xiaobin Hu, Deng-Ping Fan, Ling Shao, and Luc Van~Gool.
\newblock Highly accurate dichotomous image segmentation.
\newblock In \emph{European Conference on Computer Vision}, pages 38--56. Springer, 2022{\natexlab{b}}.

\bibitem[Quattrini et~al.(2024)Quattrini, Pippi, Cascianelli, and Cucchiara]{quattrini2024alfie}
Fabio Quattrini, Vittorio Pippi, Silvia Cascianelli, and Rita Cucchiara.
\newblock Alfie: Democratising rgba image generation with no \$\$\$.
\newblock \emph{arXiv preprint arXiv:2408.14826}, 2024.

\bibitem[Rombach et~al.(2022)Rombach, Blattmann, Lorenz, Esser, and Ommer]{latentdiffusion}
Robin Rombach, Andreas Blattmann, Dominik Lorenz, Patrick Esser, and Björn Ommer.
\newblock High-resolution image synthesis with latent diffusion models, 2022.

\bibitem[Rother et~al.(2004)Rother, Kolmogorov, and Blake]{rother2004grabcut}
Carsten Rother, Vladimir Kolmogorov, and Andrew Blake.
\newblock "grabcut": interactive foreground extraction using iterated graph cuts.
\newblock \emph{ACM transactions on graphics (TOG)}, 23\penalty0 (3):\penalty0 309--314, 2004.

\bibitem[Saharia et~al.(2022)Saharia, Chan, Saxena, Li, Whang, Denton, Ghasemipour, Ayan, Mahdavi, Lopes, Salimans, Ho, Fleet, and Norouzi]{imagen}
Chitwan Saharia, William Chan, Saurabh Saxena, Lala Li, Jay Whang, Emily Denton, Seyed Kamyar~Seyed Ghasemipour, Burcu~Karagol Ayan, S.~Sara Mahdavi, Rapha~Gontijo Lopes, Tim Salimans, Jonathan Ho, David~J Fleet, and Mohammad Norouzi.
\newblock Photorealistic text-to-image diffusion models with deep language understanding, 2022.

\bibitem[Sarukkai et~al.(2024)Sarukkai, Li, Ma, R{\'e}, and Fatahalian]{sarukkai2024collage}
Vishnu Sarukkai, Linden Li, Arden Ma, Christopher R{\'e}, and Kayvon Fatahalian.
\newblock Collage diffusion.
\newblock In \emph{Proceedings of the IEEE/CVF winter conference on applications of computer vision}, pages 4208--4217, 2024.

\bibitem[Shinn et~al.(2024)Shinn, Cassano, Gopinath, Narasimhan, and Yao]{shinn2024reflexion}
Noah Shinn, Federico Cassano, Ashwin Gopinath, Karthik Narasimhan, and Shunyu Yao.
\newblock Reflexion: Language agents with verbal reinforcement learning.
\newblock \emph{Advances in Neural Information Processing Systems}, 36, 2024.

\bibitem[Sohl-Dickstein et~al.(2015)Sohl-Dickstein, Weiss, Maheswaranathan, and Ganguli]{sohl2015deep}
Jascha Sohl-Dickstein, Eric Weiss, Niru Maheswaranathan, and Surya Ganguli.
\newblock Deep unsupervised learning using nonequilibrium thermodynamics.
\newblock In \emph{International conference on machine learning}, pages 2256--2265, 2015.

\bibitem[Song et~al.(2021)Song, Meng, and Ermon]{song2021denoising}
Jiaming Song, Chenlin Meng, and Stefano Ermon.
\newblock Denoising diffusion implicit models.
\newblock In \emph{International Conference on Learning Representations}, 2021.

\bibitem[Tang et~al.(2022)Tang, Liu, Pandey, Jiang, Yang, Kumar, Stenetorp, Lin, and Ture]{tang2022daam}
Raphael Tang, Linqing Liu, Akshat Pandey, Zhiying Jiang, Gefei Yang, Karun Kumar, Pontus Stenetorp, Jimmy Lin, and Ferhan Ture.
\newblock What the daam: Interpreting stable diffusion using cross attention.
\newblock \emph{arXiv preprint arXiv:2210.04885}, 2022.

\bibitem[Team(2024)]{omost}
Omost Team.
\newblock Omost github page.
\newblock \url{https://github.com/lllyasviel/Omost}, 2024.
\newblock Accessed: 2024-11-06.

\bibitem[Tumanyan et~al.(2023)Tumanyan, Geyer, Bagon, and Dekel]{tumanyan2023plug}
Narek Tumanyan, Michal Geyer, Shai Bagon, and Tali Dekel.
\newblock Plug-and-play diffusion features for text-driven image-to-image translation.
\newblock In \emph{Proceedings of the IEEE/CVF Conference on Computer Vision and Pattern Recognition}, pages 1921--1930, 2023.

\bibitem[vidu(2024)]{vidu}
vidu.
\newblock vidu.
\newblock \url{https://www.vidu.com/}, 2024.
\newblock Accessed: 2025-02-27.

\bibitem[Wang et~al.(2024)Wang, Li, Zhu, Guo, Dou, and Li]{wang2024customvideo}
Zhao Wang, Aoxue Li, Lingting Zhu, Yong Guo, Qi Dou, and Zhenguo Li.
\newblock Customvideo: Customizing text-to-video generation with multiple subjects.
\newblock \emph{arXiv preprint arXiv:2401.09962}, 2024.

\bibitem[Yao et~al.(2024)Yao, Wang, Yang, and Wang]{yao2024vitmatte}
Jingfeng Yao, Xinggang Wang, Shusheng Yang, and Baoyuan Wang.
\newblock Vitmatte: Boosting image matting with pre-trained plain vision transformers.
\newblock \emph{Information Fusion}, 103:\penalty0 102091, 2024.

\bibitem[Zhang and Agrawala(2024)]{layerdiffusion}
Lvmin Zhang and Maneesh Agrawala.
\newblock Transparent image layer diffusion using latent transparency.
\newblock \emph{arXiv preprint arXiv:2402.17113}, 2024.

\bibitem[Zhang et~al.(2023)Zhang, Rao, and Agrawala]{zhang2023adding}
Lvmin Zhang, Anyi Rao, and Maneesh Agrawala.
\newblock Adding conditional control to text-to-image diffusion models.
\newblock In \emph{Proceedings of the IEEE/CVF International Conference on Computer Vision}, pages 3836--3847, 2023.

\bibitem[Zhang et~al.(2024)Zhang, Song, Liu, Wang, Yu, Tang, Li, Tang, Hu, Pan, et~al.]{zhang2024ssr}
Yuxuan Zhang, Yiren Song, Jiaming Liu, Rui Wang, Jinpeng Yu, Hao Tang, Huaxia Li, Xu Tang, Yao Hu, Han Pan, et~al.
\newblock Ssr-encoder: Encoding selective subject representation for subject-driven generation.
\newblock In \emph{Proceedings of the IEEE/CVF Conference on Computer Vision and Pattern Recognition}, pages 8069--8078, 2024.

\bibitem[Zheng et~al.(2024{\natexlab{a}})Zheng, Gao, Fan, Liu, Laaksonen, Ouyang, and Sebe]{birefnet}
Peng Zheng, Dehong Gao, Deng-Ping Fan, Li Liu, Jorma Laaksonen, Wanli Ouyang, and Nicu Sebe.
\newblock Bilateral reference for high-resolution dichotomous image segmentation.
\newblock \emph{arXiv preprint arXiv:2401.03407}, 2024{\natexlab{a}}.

\bibitem[Zheng et~al.(2024{\natexlab{b}})Zheng, Gao, Fan, Liu, Laaksonen, Ouyang, and Sebe]{zheng2024bilateral}
Peng Zheng, Dehong Gao, Deng-Ping Fan, Li Liu, Jorma Laaksonen, Wanli Ouyang, and Nicu Sebe.
\newblock Bilateral reference for high-resolution dichotomous image segmentation.
\newblock \emph{arXiv preprint arXiv:2401.03407}, 2024{\natexlab{b}}.

\end{thebibliography}
}
% \input{reference.bbl}
% \bibliographystyle{IEEEtran}
% \bibliography{reference}
% \input{sec/X_suppl}
% \input{sec/rebuttal}
\end{document}